\documentclass[10pt,twocolumn,letterpaper]{article}

\usepackage[pagenumbers]{iccv} %

\usepackage{bm}
\usepackage{acro}
\usepackage{xspace}
\usepackage{array}
\usepackage{booktabs}
\usepackage{fontawesome}
\usepackage{xcolor,colortbl}
\definecolor{Gray}{gray}{0.9}
\definecolor{Textgray}{gray}{0.4}
\usepackage[normalem]{ulem}
\definecolor{notetext}{rgb}{0.7,0,0}
\definecolor{notetext}{rgb}{0.7,0,0}

\usepackage[nointegrals]{wasysym}
\usepackage{makecell}
\usepackage{balance}
\usepackage{multirow}
\usepackage{overpic}
\usepackage{bbm}
\usepackage{dsfont}
\usepackage[font=small,labelfont=bf]{caption}
\usepackage{color}
\usepackage{pifont}
\usepackage{etoolbox}
\usepackage{float}
\usepackage{tikz}
\usepackage{svg}
\usepackage{times}
\usepackage{epsfig}
\usepackage{graphicx}
\usepackage{amsmath}
\usepackage{amssymb}
\usepackage{comment}
\usepackage{soul}
\usepackage{listings}
\usepackage{lscape}
\usepackage{rotating}
\usepackage{enumitem}
\usepackage[utf8]{inputenc}
\usepackage{pgfplots}
\usepackage{import}
\usepackage{xifthen}
\usepackage{siunitx}
\usepackage{transparent}
\usepackage{tabularx}
\usepackage{tabu}
\usepackage{xcolor}
\usepackage{arydshln}
\DeclareUnicodeCharacter{2212}{−}
\usepgfplotslibrary{groupplots,dateplot}
\usetikzlibrary{patterns,shapes.arrows}
\pgfplotsset{compat=newest}

\usepackage{silence}
\usepackage{subcaption}
\usepackage{xfrac}
\usepackage{tcolorbox}

\definecolor{m_red}{RGB}{255,209,209}
\definecolor{m_red_border}{RGB}{215,23,20}

\definecolor{m_orange}{RGB}{226,213,231}
\definecolor{m_orange_border}{RGB}{150,114,164}

\definecolor{m_blue}{RGB}{217,232,251}
\definecolor{m_blue_border}{RGB}{107,141,190}

\definecolor{m_yellow}{RGB}{255,242,205}
\definecolor{m_yellow_border}{RGB}{213,182,82}

\definecolor{m_gray}{RGB}{245,245,245}
\definecolor{m_gray_border}{RGB}{102,102,102}

\newcommand{\circlenum}[1]{{\textcircled{\footnotesize{#1}}}}

\lstdefinestyle{mypython}{
  language=python,
  breaklines=true,
  basicstyle=\fontsize{8}{12}\selectfont\ttfamily,
  keywordstyle=\bfseries\color{my_blue},
  linewidth=.99\textwidth,
}

\newcommand{\PAR}[1]{\vskip4pt \noindent {\bf #1~}}

\definecolor{darkgreen}{RGB}{0,255,0}
\definecolor{linkgreen}{RGB}{52,130,48}

\newcommand{\cmark}{\ding{51}}%
\newcommand{\xmark}{\ding{55}}%

\newcommand{\JF}{$\mathcal{J}$\&$\mathcal{F}$}%

\newcolumntype{Y}{>{\centering\arraybackslash}X}
\newcolumntype{Z}{>{\raggedleft\arraybackslash}X}
\newcolumntype{s}{>{\centering}p}

\newif\ifmynotes
\mynotestrue
\definecolor{notetext}{rgb}{0.7,0,0}

\makeatletter
\def\adl@drawiv#1#2#3{%
        \hskip.5\tabcolsep
        \xleaders#3{#2.5\@tempdimb #1{1}#2.5\@tempdimb}%
                #2\z@ plus1fil minus1fil\relax
        \hskip.5\tabcolsep}
\newcommand{\cdashlinelr}[1]{%
  \noalign{\vskip\aboverulesep
           \global\let\@dashdrawstore\adl@draw
           \global\let\adl@draw\adl@drawiv}
  \cdashline{#1}
  \noalign{\global\let\adl@draw\@dashdrawstore
           \vskip\belowrulesep}}
\makeatother

\makeatletter
\newcommand\smaller{\@setfontsize\smaller{8.7}{9.5}}
\makeatother

\definecolor{iccvblue}{rgb}{0.21,0.49,0.74}
\usepackage[pagebackref,breaklinks,colorlinks,allcolors=iccvblue]{hyperref}

\title{Sa2VA-i: Improving Sa2VA Results with Consistent Training and Inference}

\author{
Alexey Nekrasov$^1$ \and Ali Athar \and Daan de Geus$^2$ \and Alexander Hermans$^1$ \and Bastian Leibe$^1$ \\
\vspace{1em}
$^1$~RWTH Aachen University \quad $^2$~Eindhoven University of Technology%
}

\begin{document}
\maketitle
\begin{abstract}
Sa2VA is a recent model for language-guided dense grounding in images and video that achieves state-of-the-art results on multiple segmentation benchmarks and that has become widely popular.
However, we found that Sa2VA does not perform according to its full potential for referring video object segmentation tasks.
We identify inconsistencies between training and inference procedures as the key factor holding it back.
To mitigate this issue, we propose an improved version of Sa2VA, Sa2VA-i, that rectifies these issues and improves the results.
In fact, Sa2VA-i sets a new state of the art for multiple video benchmarks and achieves improvements of up to +11.6 $\mathcal{J}$\&$\mathcal{F}$ on MeViS, +1.4 on Ref-YT-VOS, +3.3 on Ref-DAVIS and +4.1 on ReVOS using the same Sa2VA checkpoints.
With our fixes, the Sa2VA-i-1B model even performs on par with the original Sa2VA-26B model on the MeViS benchmark.
We hope that this work will show the importance of seemingly trivial implementation details and that it will provide valuable insights for the referring video segmentation field.
We provide the code and updated models at \url{https://github.com/kumuji/sa2va-i}.
\end{abstract}

\section{Introduction}
\label{sec:intro}

Accurately localizing and tracking objects in videos is an important area in computer vision with multiple real-world applications, ranging from video editing~\cite{bekuzarov2023xmemplus} and robotics~\cite{lu2023selfsupervised} to medical~\cite{khader2024medical} and biological~\cite{yang2024deep} applications.
Over the years, a large volume of work~\cite{xu2018youtubevos,yang2019youtubevis,qi2022ovis,miao2022vipseg,athar2023burst,perazzi2016davis} proposed datasets, benchmarks, and model architectures that focus on segmenting and tracking objects.
The recent surge in vision-language models~\cite{liu2023llava,alayrac2022flamingo} has transformed the landscape of video segmentation research, particularly in segmenting and tracking objects based on complex text descriptions.
This task, called Referring (and later extended to Reasoning) Video Object Segmentation (RVOS), has recently attracted significant attention, with multiple new datasets~\cite{ding2023mevis,bellver2020refvos,yan2024visa,athar2025vicas} and methods~\cite{bai2024videolisa,yan2024visa,yuan2025sa2va} emerging.

Earlier architectures that addressed RVOS~\cite{botach2022mttr,wu2022referformer} relied on vision-language models such as CLIP~\cite{radford2021clip} to extract features from the image frames and text prompts, coupled with a segmentation model that predicted masks.
While effective for short text prompts, this approach struggles with detailed text descriptions requiring logical reasoning.
Therefore, recent works~\cite{xia2024gsva,bai2024videolisa,lai2023lisa,yan2024visa,munasinghe2024videoglamma} employ Multimodal Large Language Models (MLLM), which are given video features and a text prompt as input and provide segmentation masks as an output.
The prevailing practice is to train the MLLM to predict a special \texttt{[SEG]} token that represents the object to be segmented, and to use a segmentation model such as SAM or SAM2~\cite{kirillov2023sam,ravi2025sam2} to predict a segmentation mask based on this token. %
The current state-of-the-art model, Sa2VA~\cite{yuan2025sa2va}, adopts this paradigm, but scales the approach with more data and uses more recent MLLMs.

Sa2VA introduces a prompting strategy that uses the streaming memory of SAM2 to segment objects in a video.
Despite achieving great results on multiple benchmarks, we identify a critical discrepancy between the training and inference procedure.
During training, only the SAM2 mask decoder is finetuned, while the memory encoder and attention components remain frozen and unused.
This is inconsistent with the inference procedure, because these frozen components \textit{are} used during inference, operating on feature representations from a finetuned mask decoder, which they were never optimized to handle.
Thus, although the memory prompting mechanism is theoretically sound, its performance is suboptimal due to this mismatch between training and inference, particularly on challenging video segmentation benchmarks such as MeViS and ReVOS~\cite{ding2023mevis,yan2024visa}.

In this paper, we address this limitation by proposing Sa2VA-i, an improved variant of Sa2VA.
To mitigate the aforementioned inconsistency, we ensure that the model's operation during inference is the same as during training.
Concretely, this means that Sa2VA-i predicts segmentation masks \textit{per frame}, without using the memory encoder or attention, during both training and inference.
To allow the model to make predictions on long videos, Sa2VA-i post-processes the initial per-frame mask predictions with an off-the-shelf, non-finetuned SAM2 decoder, for which we store the weights in the same released checkpoint.
Additionally, to further improve performance, we revisit several other design choices related to frame sampling during inference.
Our experiments show that Sa2VA-i significantly outperforms the original model, sets a new state of the art for referring and reasoning segmentation among published methods, and achieves the 3rd place on the test set of LSVOS 20205 MeViS Track.
We hope that our detailed analysis of these implementation nuances will guide future research on improving video segmentation architectures.
We release the updated models and code which provide drop-in replacements for the original pipeline.

\section{Related Work}
\label{sec:relwork}

\PAR{Video Segmentation with MLLMs}
With recent advances in vision-language research, multiple referring image segmentation methods have been proposed~\cite{xia2024gsva,lai2023lisa,zhang2024psalm,rasheed2024glamm}, sparking interest in segmenting and tracking objects in \textit{videos}, \ie, Reasoning and Referring Video Object Segmentation (RVOS).
Although initial architectures for RVOS involved using text and vision encoders coupled with a Transformer decoder to regress masks~\cite{yan2023mutr,botach2022mttr,wu2022referformer,bellver2020refvos,seo2020urvos}, recent approaches~\cite{bai2024videolisa,yan2024visa,yuan2025sa2va} adopt MLLM-based architectures to better reason about complex text prompts and to combine multiple tasks.
This is usually done by extending a Multimodal LLM architecture with a downstream segmentation model and by predicting a special segmentation token for each target object.
This token is then input into the segmentation model together with video frame features to predict masks.
For the video segmentation task, the image-first PSALM model~\cite{zhang2024psalm} can be conditioned on a previous mask to segment the next frame.
Similarly, TrackGPT~\cite{zhu2023trackgpt} processes a single frame at a time.
VISA~\cite{yan2024visa} predicts a single token for a carefully selected keyframe in a video and tracks that corresponding object.
VideoLISA~\cite{bai2024videolisa} uses SAM~\cite{kirillov2023sam} and reuses a single token for multiple frames. %
More recently, VideoGLaMM~\cite{munasinghe2024videoglamma} and Sa2VA~\cite{yuan2025sa2va} use the SAM2~\cite{ravi2025sam2} segmentation model, but do not finetune the memory components.
In this work, we identify that Sa2VA operates differently during inference and training, which harms the performance.
To solve this, we present Sa2VA-i, which ensures consistent training and inference and considerably outperforms Sa2VA.

\PAR{Works that use Sa2VA}
Sa2VA has rapidly become a foundational model for language-grounded segmentation, and has been widely adopted and extended across diverse research efforts.
For instance, its capabilities are leveraged for dataset creation, with AnimeShooter~\cite{qiu2025animeshooter} using Sa2VA to extract masks of animated characters.
Furthermore, several new benchmarks, including EASG-Bench~\cite{rodin2025easgbench}, FinChart-Bench~\cite{shu2025finchartbench} and V-STaR~\cite{cheng2025vstar}, evaluate Sa2VA as a state-of-the-art method for spatial grounding.
The model is also a baseline for comparison to SpatialReasoner-R1~\cite{shen2025finegrained}, Omni-R1~\cite{zhong2025omnir1}, SeC~\cite{zhang2025sec}, and DenseWorld-1M~\cite{li2025denseworld1m}, EgoMask~\cite{liang2025egomask}.
InterRVOS~\cite{jin2025interrvos}, Dense360~\cite{zhou2025dense360}, UniBiomed~\cite{wu2025unibiomed}, DeSa2VA~\cite{jisheng2025desa2va}, and SAMA~\cite{sun2025sama} are built directly on top of Sa2VA.
The practical utility of Sa2VA is evident in applications such as Gondola~\cite{chen2025gondola} and FineGrasp~\cite{du2025finegrasp}, which use it to segment objects for grasping.
Furthermore, Sa2VA prominence was highlighted during the PVUW MeViS Challenge at CVPR 2025~\cite{ding2024pvuw}, where the first~\cite{fang2025pvuw1st} and third~\cite{yuan2025pvuw3rd} place solutions were based on Sa2VA.
The frequent use of Sa2VA in the literature highlights the relevance of this work in which we identify and solve the inference-training inconsistency in Sa2VA.
With Sa2VA-i, we provide a stronger base model for various benchmarks, applications and downstream tasks.
We encourage authors to apply our suggested improvements to ensure proper comparisons and improve their results.

\section{Method}

\subsection{Preliminaries}

\PAR{SAM and SAM2}
SAM~\cite{kirillov2023sam} is an interactive image segmentation model that consists of image encoder, prompt encoder and mask decoder.
Box, point or mask inputs from the user are encoded by the prompt encoder into a single feature vector.
Using this feature vector, as well as image features encoded by the image encoder, the mask decoder generates a segmentation output for the selected object.
For interactive video segmentation, SAM2~\cite{ravi2025sam2} augments the original SAM architecture with several \textit{memory} components.
SAM2 processes videos in a streaming fashion, \ie, frame by frame.
Like SAM, SAM2 segments objects based on a box prompt, mask prompt or point prompt, which are provided for one of the video frames in this case.
Given an input frame, SAM2 first generates per-frame features with its image encoder.
Then, SAM2 applies \textit{memory attention} to condition these per-frame features on the features from earlier frames in which an object was segmented, which were stored in a \textit{memory bank}.
Subsequently, the \textit{mask decoder} predicts a segmentation mask based on the conditioned frame and the input prompts.
Finally, the image and object features are encoded using a \textit{memory encoder} and stored in the memory bank such that they can be used to condition future frames.

\PAR{Sa2VA}
To conduct referring segmentation, Sa2VA combines a Multimodal Large Language Model (MLLM) with SAM2.
First, it feeds the video frames through the vision encoder of the MLLM to obtain visual tokens, which are then concatenated with the input text tokens.
These tokens are the input to the MLLM, which jointly processes the data and predicts a sequence of output text tokens.
To obtain segmentation masks, a special \texttt{[SEG]} text token is added to the vocabulary, which the MLLM has to generate when it wants to output a segmentation mask.
To turn the \texttt{[SEG]} token into a segmentation mask, this token's latent feature vector is obtained from the final MLLM layer, passed to an MLP, and finally passed to SAM2 as an object prompt, along with the corresponding image frames.

For referring \textit{image} segmentation, the SAM2 model effectively acts like a SAM model.
The feature vector of the \texttt{[SEG]} token circumvents the prompt encoder and is directly passed to the mask decoder to predict the final mask.
In this case, no memory components are used to make a prediction, as there is only one input frame.

For referring \textit{video} segmentation, Sa2VA also predicts only a single \texttt{[SEG]} token, and reuses this token to predict masks for multiple video frames, similar to VideoLISA~\cite{bai2024videolisa}.
Due to the limited context window of the MLLM, Sa2VA samples $T$ video frames, only feeds the visual tokens for those $T$ frames through the MLLM, and makes segmentation predictions with SAM2 only for those frames.
To obtain predictions for all $I>T$ frames in a video, Sa2VA applies post-processing using the same SAM2 decoder.

\PAR{Sa2VA Video Segmentation Training}
During training, Sa2VA uses SAM2 to predict segmentation masks for each video frame \textit{separately}, using the same reused \texttt{[SEG]} token and without using any of the memory components.
Concretely, for each input video frame, the predicted \texttt{[SEG]} token's features and the video frame features are directly fed into the mask decoder to predict a segmentation mask, without any conditioning on memory features from earlier frames.
This effectively reduces SAM2 to SAM, and means that only SAM2's mask decoder weights are updated when training the model, leaving the memory components untouched.

To supervise the MLLM's output tokens, Sa2VA uses the cross-entropy loss that is standard for next-token prediction.
For the segmentation masks, Sa2VA is optimized using a combination of the Dice loss and the cross-entropy loss.

\begin{figure}[t]
    \centering
    \includegraphics[width=\linewidth]{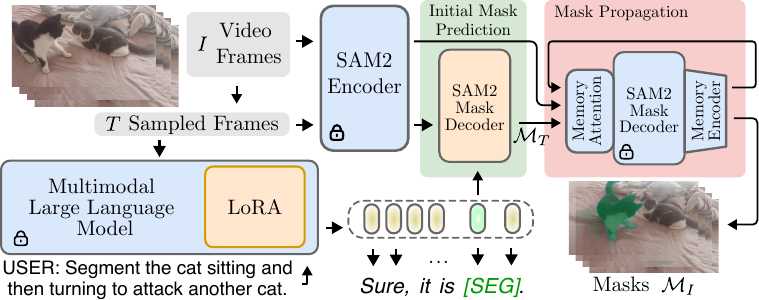}
    \caption{
    \textbf{Sa2VA-i Architecture Overview.}
    Sa2VA-i first predicts initial masks $\mathcal{M}_T$ using a \textit{finetuned} SAM2 mask decoder by taking the generated \texttt{[SEG]} token and predicting a mask for all $T$ sampled frames separately. Next, it propagates these  $\mathcal{M}_T$ masks across all video frames $I$ using SAM2's \textit{original} mask decoder, yielding output masks $\mathcal{M}_I$.
    }
    \label{fig:mainfig}
\end{figure}

\PAR{Sa2VA Video Segmentation Inference}
The main issue arises during inference, where the model's operation diverges significantly from its operation during training.
Concretely, during inference, Sa2VA \textit{does} use the memory components of SAM2, and propagates masks in a streaming fashion.
Sa2VA uses the predicted \texttt{[SEG]} token and the memory encoder to initialize the memory bank, and then conditions each next frame on the memory bank using memory attention, propagating the segmentation mask through the video and making a prediction for each frame.
However, the memory components used by the model -- \ie, the memory encoder and memory attention -- \emph{were not finetuned} during training.
This means that their weights were not trained to handle the features generated by the mask decoder that \emph{was finetuned} during training.
And vice versa, the finetuned mask decoder is only fed unconditioned frame features during training, and cannot properly handle conditioned features during inference as a result.
This incompatibility between the frozen memory components and the finetuned mask decoder harms the performance of the model.

\begin{table*}[ht]
\centering
\renewcommand{\tabcolsep}{5pt}
\caption{
\textbf{Comparison to State of the Art.}
We observe consistent improvements to Sa2VA across all popular video segmentation benchmarks, and obtain new state-of-the-art performance.
We observe that the optimal number of sampled frames during testing depends on the dataset, where more complex benchmarks benefit from a larger number of sampled frames.
\textit{FT} Denotes a model trained with 8 video frames and uniform sampling during training.
}
\label{tab:sota_mevis_ytvos_davis}
\begin{tabularx}{1.0\textwidth}{llc YYY c YYY c YYY}
\toprule
\multirow{2}[2]{*}{Method} & \multirow{2}[2]{*}{MLLM} & \multirow{2}[2]{*}{Frames} &
\multicolumn{3}{c}{MeViS~\cite{ding2023mevis}} &&
\multicolumn{3}{c}{Ref-YT-VOS~\cite{seo2020urvos}} &&
\multicolumn{3}{c}{Ref-DAVIS17~\cite{khoreva2018rdavis}} \\
\cmidrule{4-6} \cmidrule{8-10} \cmidrule{12-14}
& & &
$\mathcal{J}\&\mathcal{F}$ & $\mathcal{J}$ & $\mathcal{F}$ &&
$\mathcal{J}\&\mathcal{F}$ & $\mathcal{J}$ & $\mathcal{F}$ &&
$\mathcal{J}\&\mathcal{F}$ & $\mathcal{J}$ & $\mathcal{F}$ \\

\midrule

LISA~\cite{lai2023lisa} & LLaVA-7B & -- &
37.2 & 35.1 & 39.4 &&
53.9 & 53.4 & 54.3 &&
64.8 & 62.2 & 67.3 \\

LISA~\cite{lai2023lisa} & LLaVA-13B & -- &
37.9 & 35.8 & 40.0 &&
54.4 & 54.0 & 54.8 &&
66.0 & 63.2 & 68.8 \\

TrackGPT~\cite{zhu2023trackgpt} & LLaVA-7B & -- &
40.1 & 37.6 & 42.6 &&
56.4 & 55.3 & 57.4 &&
63.2 & 59.4 & 67.0 \\

TrackGPT~\cite{zhu2023trackgpt} & LLaVA-13B & -- &
41.2 & 39.2 & 43.1 &&
59.5 & 58.1 & 60.8 &&
66.5 & 62.7 & 70.4 \\

PSALM~\cite{zhang2024psalm} & Phi-1.5 & -- &
--- & --- & --- &&
--- & --- & --- &&
68.8 & 65.9 & 71.7 \\

VISA~\cite{yan2024visa} & Chat-UniVi-7B & -- &
43.5 & 40.7 & 46.3 &&
61.5 & 59.8 & 63.2 &&
69.4 & 66.3 & 72.5 \\

VISA~\cite{yan2024visa} & Chat-UniVi-13B & -- &
44.5 & 41.8 & 47.1 &&
63.0 & 61.4 & 64.7 &&
70.4 & 67.0 & 73.8\\

VideoLISA~\cite{bai2024videolisa} & LLaVa-Phi-3-V-3.8B & -- &
44.4 & 41.3 & 47.6 &&
63.7 & 61.7 & 65.7 &&
68.8 & 64.9 & 72.7\\

InstructSeg~\cite{wei2024instructseg} & Mipha-3B &&
--- & --- & --- &&
67.5 & 65.4 & 69.5 &&
71.1 & 67.3 & 74.9 \\

VideoGLaMM~\cite{munasinghe2024videoglamma} & Phi3-Mini-3.8B & -- &
45.2 & 42.0 & 48.2 &&
--- & --- & --- &&
--- & --- & --- \\

SAMWISE~\cite{cuttano2024samwisea} & RoBERTa & -- &
49.5 & 46.6 & 52.4 &&
69.2 & 67.8 & 70.6 &&
70.6 & 67.4 & 74.5 \\

VRS-HQ~\cite{gong2025vrshq} & Chat-UniVi-13B & -- &
50.9 & 48.0 & 53.7 &&
71.0 & 69.0 & 73.1 &&
74.4 & 71.0 & 77.9 \\

GLUS~\cite{lin2025glus} & LLaVA-7B & -- &
51.3 & 48.5 & 54.2 && 
67.3 & 65.5 & 69.0 &&
-- & -- & -- \\

MPG-SAM 2~\cite{rong2025mpgsam} & BEiT & -- &
53.7 & 50.7 & 56.7 &&
73.9 & 71.7 & 76.1 &&
72.4 & 68.8 & 76.0 \\

\midrule
Sa2VA~\cite{yuan2025sa2va} & InternVL2.5-1B & 5 &
47.0 & -- & -- &&
68.0 & -- & -- &&
69.5 & -- & -- \\
Sa2VA~\cite{yuan2025sa2va} & InternVL2.5-4B & 5 &
46.4 & 43.3 & 49.5 &&
71.3 & 69.1 & 73.5 &&
73.7 & 69.6 & 77.8 \\
Sa2VA~\cite{yuan2025sa2va} & InternVL2.5-8B & 5 &
51.5 & -- & -- &&
72.3 & -- & -- &&
75.9 & -- & -- \\
Sa2VA~\cite{yuan2025sa2va} & InternVL2.5-26B & 5 &
52.1 & -- & -- &&
75.1 & -- & -- &&
78.6 & -- & -- \\

\midrule

Sa2VA-i & InternVL2.5-1B & 5 &
52.2 & 49.6 & 54.9 &&
70.9 & 68.9 & 72.8 &&
74.5 & 70.9 & 78.0 \\
Sa2VA-i & InternVL2.5-4B & 5 &
56.2 & 53.5 & 59.0 &&
74.1 & 71.2 & 76.3 &&
77.6 & 74.0 & 81.1 \\
Sa2VA-i & InternVL2.5-8B & 5 &
58.7 & 55.7 & 58.7 &&
74.7 & 72.5 & 76.7 &&
80.1 & 76.7 & 83.5 \\
Sa2VA-i & InternVL2.5-26B & 5 &
62.2 & 59.2 & 65.2 &&
\ul{76.4} & \ul{74.2} & \ul{78.6} &&
\textbf{81.9} & \textbf{78.4} & \textbf{85.4} \\

\midrule

Sa2VA-i & InternVL2.5-1B & 8 &
52.6 & 49.9 & 55.3 &&
70.3 & 68.4 & 72.2 &&
73.6 & 70.1 & 77.1 \\
Sa2VA-i & InternVL2.5-4B & 8 &
56.6 & 53.9 & 59.3 &&
73.2 & 71.0 & 75.4 &&
78.6 & 75.1 & 82.1 \\
Sa2VA-i & InternVL2.5-8B & 8 &
59.5 & 56.6 & 62.4 &&
73.9 & 71.7 & 76.3 &&
79.1 & 75.6 & 82.7 \\
Sa2VA-i & InternVL2.5-26B & 8 &
\ul{63.2} & \ul{60.1} & \ul{66.2} &&
\textbf{76.5} & \textbf{74.3} & \textbf{78.7} &&
\ul{81.2} & \ul{77.5} & \ul{84.9} \\

\midrule

Sa2VA-i-FT & InternVL2.5-4B & 8 &
57.3 & 54.6 & 60.0 &&
74.1 & 71.9 & 76.3 &&
79.3 & 75.8 & 82.7 \\

\midrule

Sa2VA-i & InternVL2.5-26B & 12 &
\textbf{63.7} & \textbf{60.6} & \textbf{66.8} &&
75.7 & 73.5 & 77.9 &&
80.7 & 77.0 & 84.3 \\

\bottomrule
\end{tabularx}
\end{table*}

\begin{table*}[t]
\centering
\renewcommand{\tabcolsep}{4pt}
\caption{
\textbf{Reasoning Video Segmentation on ReVOS~\cite{yan2024visa}.}
Sa2VA-i outperforms Sa2VA and other methods by a large margin, obtaining state-of-the-art performance.
}
\label{tab:sota_revos}
\begin{tabularx}{1.0\linewidth}{ll c YYY c YYY c YYY}
\toprule
\multirow{2}[2]{*}{Method} & \multirow{2}[2]{*}{MLLM} &&
\multicolumn{3}{c}{Reasoning} &&
\multicolumn{3}{c}{Referring} &&
\multicolumn{3}{c}{Overall} \\

\cmidrule{4-6} \cmidrule{8-10} \cmidrule{12-14}

& & & 
$\mathcal{J}\&\mathcal{F}$ & $\mathcal{J}$ & $\mathcal{F}$ &&
$\mathcal{J}\&\mathcal{F}$ & $\mathcal{J}$ & $\mathcal{F}$ &&
$\mathcal{J}\&\mathcal{F}$ & $\mathcal{J}$ & $\mathcal{F}$ \\

\midrule

VISA~\cite{yan2024visa} & Chat-UniVi-7B && 
39.2 & 36.7 & 41.7 && 
52.9 & 51.1 & 54.7 && 
46.1 & 43.9 & 48.2 \\

VISA~\cite{yan2024visa} & Chat-UniVi-13B && 
40.9 & 38.3 & 43.5 && 
54.1 & 52.3 & 55.8 && 
47.5 & 45.3 & 49.7 \\

InstructSeg~\cite{wei2024instructseg} & Mipha-3B && 
51.9 & 49.2 & 54.7 && 
57.0 & 54.8 & 59.2 && 
54.5 & 52.0 & 56.9 \\

VRS-HQ~\cite{gong2025vrshq} & Chat-UniVi-13B && 
56.8 & 54.1 & 59.4 &&
63.3 & 61.1 & 65.5 && 
60.0 & 57.6 & 62.5  \\

Sa2VA~\cite{yuan2025sa2va} & InternVL2.5-4B &&
56.2 & 53.2 & 59.1 &&
63.0 & 60.5 & 65.5 &&
59.6 & 56.8 & 62.3 \\

\midrule

Sa2VA-i & InternVL2.5-1B && 
48.3 & 45.9 & 50.7 && 
58.6 & 56.4 & 60.7 && 
53.4 & 51.2 & 50.7 \\

Sa2VA-i & InternVL2.5-4B && 
60.9 & 58.2 & 63.6 && 
66.5 & 64.3 & 68.8 && 
63.7 & 61.2 & 66.1 \\

Sa2VA-i & InternVL2.5-8B && 
\ul{63.1} & \ul{60.3} & \ul{65.8} && 
\ul{68.7} & \ul{66.3} & \ul{71.1} && 
\ul{65.9} & \ul{63.3} & \ul{68.5} \\

Sa2VA-i & InternVL2.5-26B && 
\textbf{65.8} & \textbf{62.9} & \textbf{71.3} && 
\textbf{71.3} & \textbf{69.0} & \textbf{73.5} && 
\textbf{68.5} & \textbf{65.9} & \textbf{71.0} \\

\midrule 

Sa2VA-i-FT & InternVL2.5-4B && 
61.5 & 58.8 & 64.2 && 
67.1 & 64.8 & 69.4 && 
64.3 & 61.8 & 66.8 \\

\bottomrule
\end{tabularx}
\label{tab:reason}
\end{table*}

\subsection{Sa2VA-i}
Our model, called Sa2VA-i and shown in \cref{fig:mainfig}, addresses this problem by ensuring the inference procedure is consistent with the training procedure.
To achieve this, during inference, we do \textit{not} use SAM2's memory components to predict segmentation masks and follow the exact same procedure that is followed during training.

Concretely, we take the predicted \texttt{[SEG]} token's features and the per-frame video features and feed them directly to the mask decoder, to predict segmentation masks $\mathcal{M}_T$ for the $T$ sampled frames.
Subsequently, to make a prediction for all $I$ frames of the video, we use an off-the-shelf, \textit{non-finetuned} SAM2 decoder weights.
Concretely, we directly prompt SAM2 with predicted masks $\mathcal{M}_T$, and use the original SAM2 inference procedure to predict masks $\mathcal{M}_I$ for all frames.
By following this approach, there is no longer any incompatibility between non-finetuned memory components and a finetuned mask decoder, because (a) no memory components are used to make the initial predictions $\mathcal{M}_T$, and (b) the SAM2 components used to obtain the final masks $\mathcal{M}_I$ are all original and thus compatible.

In practice, this means that we have to store weights for two versions of the mask decoder: (a) the original one from SAM2, and (b) the finetuned one from Sa2VA.
The other components from SAM2 remain frozen when training Sa2VA, so the same weights can be used for initial mask prediction with Sa2VA-i and mask propagation with SAM2.
This means that there is only a small additional memory footprint of $\sim$16MB.

Additionally, we observe that the original Sa2VA uses random frame sampling during training, but samples the first $T$ video frames during inference.
This is suboptimal due to the offline nature of RVOS -- with prompts like ``the dog that disappears from the left, then re-appears'' -- where full videos have to be available during inference to answer the question properly.
Therefore, we propose to apply uniform frame sampling during inference instead.
Furthermore, to ensure further consistency between training and inference, we also train a version of Sa2VA-i that applies uniform frame sampling during training as well.

We find that these simple improvements significantly improve performance, as we will demonstrate in the next section.

\section{Experiments}
\subsection{Experimental Setup}
\PAR{Evaluation}
We evaluate Sa2VA-i using multiple popular benchmarks for video-level referring and reasoning segmentation~\cite{yan2024visa,khoreva2018rdavis,ding2023mevis,seo2020urvos}.
We use the standard $\mathcal{J}\&\mathcal{F}$ metric, which is the geometric mean of the Jaccard Score, $\mathcal{J}$, and the Boundary F-Score, $\mathcal{F}$~\cite{perazzi2016davis}.
We use the MeViS dataset~\cite{ding2023mevis} for ablations, and report scores obtained from the public evaluation server on the \texttt{val} split.

\PAR{Training}
For all of the main experiments, we use the original Sa2VA model weights. However, for an ablation experiment on the frame sampling procedure we train two 4B models ourselves, with $T=8$ video frames instead of the 5 frames used in the original Sa2VA paper, which we denote as Sa2VA-i-FT.
The training data recipe is the same as used for Sa2VA.
We use 64 A100 GPUs, each with 40GB VRAM, and training runs take approximately 8 hours.

\subsection{Comparison to State of the Art}
We compare Sa2VA-i with existing state-of-the-art referring video segmentation methods on multiple benchmarks.
As shown in \cref{tab:sota_mevis_ytvos_davis} and \cref{tab:sota_revos}, Sa2VA-i achieves state-of-the-art results on all reported benchmarks and improves upon Sa2VA results using the same checkpoints and the same number of frames.
Sa2VA-i's performance improvement is particularly substantial on the challenging MeVIS benchmark~\cite{ding2023mevis}.
It scores +10.2 \JF{} higher for the 4B model, +8.0 \JF{} for the 8B model, and +11.1 \JF{} for the 26B model.
We attribute performance improvements to the improved consistency between training and inference, as well as better frame sampling.
Additionally, we observe that the number of sampled frames during testing has a large influence on the results, but that the optimal value depends on the dataset.

\subsection{Experiments}

\begin{table}[t]
    \centering
    \caption{
    \textbf{Consistent Training and Inference.}
    Using Sa2VA's original model weights, we show gradual improvements that come from using our improved Sa2VA-i version with consistent training and inference.
    We further improve performance by using uniform sampling during inference instead of sampling first frames. 
    Evaluation on MeViS val~\cite{ding2023mevis}.
    }
    \label{tab:ablation_consistency}
    \renewcommand{\tabcolsep}{4pt}
    \begin{tabularx}{\linewidth}{m{2.2cm}m{1.4cm}m{1.2cm}YYY}
    \toprule
    Method & 
    Uniform Sampling& 
    Sampled Frames & 
    $\mathcal{J\&F}$ & 
    $\mathcal{J}$ & 
    $\mathcal{F}$ \\
    \midrule
    Sa2VA-4B~\cite{yuan2025sa2va}   & \xmark &  5 & 46.4 & 43.3 & 49.5 \\
    Sa2VA-i-4B & \xmark &  5 & 52.0 & 49.2 & 54.9 \\
    Sa2VA-i-4B & \cmark &  5 & 56.2 & 53.5 & 59.0 \\
    \bottomrule
    \end{tabularx}
\end{table}

\PAR{Consistent Training and Inference}
In \cref{tab:ablation_consistency}, we isolate the impact of (a) ensuring consistent training and inference, and (b) using uniform frame sampling.
We find that changing just the inference procedure to be consistent with the training procedure results in a substantial improvement of +5.6 \JF{} (row 1 \vs 2).
This shows the importance of solving this inconsistency.
With uniform sampling during inference, we can obtain an additional +4.2 \JF{}, bringing the total improvement to +9.8 \JF{} (row 1 \vs 3).

\begin{table}[t]
    \centering
    \caption{\textbf{Frame Sampling.} 
    We evaluate the impact of different frame sampling strategies for both Sa2VA and Sa2VA-i. Uniform sampling consistently improves performance, with the largest gains observed when training and inference sampling are matched.
    Evaluation on MeViS val~\cite{ding2023mevis}.
    }
    \label{tab:ablation_sampling}
    \renewcommand{\tabcolsep}{3pt}
    \begin{tabularx}{\linewidth}{p{2.4cm}p{1.2cm}p{1.2cm}YYY}
    \toprule
    \multirow{2}{*}{Method} &
    \multicolumn{2}{c}{Sampling} &
    \multirow{2}{*}{$\mathcal{J\&F}$} &
    \multirow{2}{*}{$\mathcal{J}$} &
    \multirow{2}{*}{$\mathcal{F}$} \\
    \cmidrule(lr){2-3}
    & Train & Test & & & \\
    \midrule
         Sa2VA-4B-FT & Random & First & 47.1 & 43.8 & 50.4 \\ 
         Sa2VA-4B-FT & Random & Uniform & 48.4 & 45.1 & 51.7 \\ 
         Sa2VA-4B-FT & Uniform & First & 48.3 & 45.2 & 51.4 \\
         Sa2VA-4B-FT & Uniform & Uniform & 51.4 & 48.3 & 54.5 \\ 
    \midrule
         Sa2VA-i-4B-FT & Random  & First & 54.6 & 51.7 & 57.5 \\ 
         Sa2VA-i-4B-FT & Random  & Uniform & \ul{56.9} & \ul{54.2} & \ul{59.7} \\ 
         Sa2VA-i-4B-FT & Uniform & First & 53.7 & 50.8 & 56.6 \\ 
         Sa2VA-i-4B-FT & Uniform & Uniform & \textbf{57.3} & \textbf{54.6} & \textbf{60.0} \\
    \bottomrule
    \end{tabularx}
\end{table}

\PAR{Frame Sampling}
We experiment with different frame sampling strategies during both training and inference, and report the results in \cref{tab:ablation_sampling}.
Similar to other works~\cite{fang2025pvuw1st,yuan2025pvuw3rd}, we observe improved performance when using uniform sampling during inference.
To ensure training-inference consistency and to see the possible gains from training the model with uniform sampling, we train a 4B-parameter model with 8 frames sampled randomly and uniformly.
During training with uniform sampling, we apply a random offset from the start to increase variation in the frames seen during training.
We observe that uniform sampling and consistent sampling strategy yields even better performance, even for the original Sa2VA model.

\begin{table}[t]
    \centering
    \caption{
    \textbf{Number of Sampled Frames.}
    Using Sa2VA's original model weights, we show that gradual improvements can be obtained by using more frames during inference.
    Performance is progressively improved, with diminishing returns when using more then 12 frames.
    Evaluation on MeViS val~\cite{ding2023mevis}.
    }
    \label{tab:ablation_num_frames}
    \renewcommand{\tabcolsep}{4pt}
    \begin{tabularx}{\linewidth}{m{2.3cm}m{1.5cm}YYY}
    \toprule
    Method & 
    Sampled Frames & 
    $\mathcal{J\&F}$ & 
    $\mathcal{J}$ & 
    $\mathcal{F}$ \\
    \midrule
    Sa2VA-i-4B &  5 & 56.2 & 53.5 & 59.0 \\
    Sa2VA-i-4B &  8 & 56.6 & 53.9 & 59.3 \\ 
    Sa2VA-i-4B & 12 & \textbf{57.4} & \textbf{54.7} & \textbf{60.0} \\ 
    Sa2VA-i-4B & 16 & \ul{57.1} & \ul{54.3} & \ul{59.9} \\ 
    Sa2VA-i-4B & 20 & 56.6 & 53.8 & 59.4 \\ 
    Sa2VA-i-4B & 24 & 56.4 & 53.6 & 59.2 \\
    \midrule
    Sa2VA-i-26B & 8 & 63.2 & 60.2 & 66.2 \\ 
    Sa2VA-i-26B & 12 & \textbf{63.7} & \textbf{60.6} & \textbf{66.8} \\ 
    Sa2VA-i-26B & 16 & \ul{63.6} & \ul{60.4} & \ul{66.8} \\ 
    Sa2VA-i-26B & 20 & 63.1 & 59.9 & 66.3 \\ 
    \bottomrule
    \end{tabularx}
\end{table}

\PAR{Number of Sampled Frames}
\cref{tab:ablation_num_frames} evaluates the impact of increasing the number of sampled frames that are fed to the MLLM and used to make the initial segmentation predictions, with the same model that has seen 5 frames during training.
We find that increasing the number of sampled frames improves performance by non-negligible margins.
In particular, the MeVIS performance can be improved by +1.2 \JF{} when using 12 sampled frames instead of 5.
These findings are similar to the results reported in~\cite{fang2025pvuw1st}, where Sa2VA obtains higher scores when using more frames.
However, while the original Sa2VA appears to benefit from using more frames without performance degradation, we observe diminishing returns for Sa2VA-i after sampling more than 12 frames.

\PAR{Overprompting of SAM2}
\begin{table}[t]
    \centering
    \caption{
    \textbf{Overprompting on Ref-DAVIS17~\cite{khoreva2018rdavis}.}
    We vary the number of frames processed by the MLLM and the number of initially predicted masks used to prompt SAM2 for mask propagation.
    On this dataset, where the target is typically visible in the first frame, performance improves as the MLLM observes more frames, while prompting SAM2 with multiple initial masks degrades performance.
    }
    \label{tab:ablation_oversampling}
    \renewcommand{\tabcolsep}{4pt}
    \begin{tabularx}{\linewidth}{p{2.3cm}p{1cm}p{1cm}YYY}
    \toprule
    \multirow{2}{*}{Method} &
    \multicolumn{2}{c}{Frames} &
    \multirow{2}{*}{$\mathcal{J\&F}$} &
    \multirow{2}{*}{$\mathcal{J}$} &
    \multirow{2}{*}{$\mathcal{F}$} \\
    \cmidrule(lr){2-3}
    & MLLM & SAM2 & & & \\
    \midrule
    Sa2VA-i-26B & 5 & 5 & \ul{81.9} & 78.4 & \ul{85.4} \\
    Sa2VA-i-26B & 5 & 1 & 81.8 & \ul{78.5} & 85.1 \\
    \midrule
    Sa2VA-i-26B & 12 & 12 & 80.7 & 77.0 & 84.3 \\
    Sa2VA-i-26B & 12 & 1 & \textbf{82.2} & \textbf{78.8} & \textbf{85.1} \\
    \bottomrule
    \end{tabularx}
\end{table}

We hypothesize that the observed diminishing returns occur partly because we ``overprompt'' the off-the-shelf SAM2 that is used for mask propagation.
In other words, we feed SAM2 with too many initial masks predicted by the MLLM and finetuned mask decoder using the single \texttt{[SEG]} token.
This may be harmful because these initial predictions by the MLLM are not perfect and can even conflict with each other, complicating SAM2's task of propagating these initial masks through all frames.
To verify our hypothesis, we conduct an experiment where we only feed the MLLM's predicted mask for the first frame to the SAM2 mask propagator, instead of feeding all $T$ uniformly sampled ones.
We select the Ref-DAVIS17 benchmark for this experiment, because the expressions in this dataset mostly refer to objects in the first frame, so the dataset benefits the least from sampling more frames (see \cref{tab:sota_mevis_ytvos_davis}).
In \cref{tab:ablation_oversampling}, we find that the performance increases from $80.7$ \JF{} to $82.2$ \JF{} when using only the first mask to prompt the SAM2 mask propagator, instead of all 12.
This confirms our hypothesis that, for this dataset, the 12 initial masks used to prompt the SAM2 mask propagator do not contain more useful information than the single mask for the first frame, and even harm the performance of the mask propagator.
However, for more complex datasets like MeViS, we observe that feeding more frames to the SAM2 mask propagator \textit{is} beneficial, as objects may appear at different moments in a video and are simply not present in some frames.
To highlight these differences between benchmarks, we encourage the community to evaluate and report scaling with the number of sampled frames for different datasets separately.

\PAR{RVOS Track of the 7th LSVOS Challenge}
\begin{table}[t]
    \centering
    \caption{
    \textbf{LSVOS 2025 MeViS Track}.
    We participate in the 7th LSVOS challenge on Referring Video Object Segmentation track and achieve the 3rd place.
    }
    \label{tab:challenge}
    \renewcommand{\tabcolsep}{4pt}
    \begin{tabularx}{\linewidth}{clYYY}
    \toprule
    Place &
    Method &
    $\mathcal{J\&F}$ &
    $\mathcal{J}$ &
    $\mathcal{F}$ \\
    \midrule
    \circlenum{1} & \texttt{niuqz} & 67.3 & 63.8 & 70.8 \\
    \circlenum{2} & \texttt{Ranhong} & 64.7 & 61.3 & 68.0 \\
    \circlenum{3} & \texttt{dytino} (Sa2VA-i) & 64.1 & 61.1 & 67.2 \\
    \bottomrule
    \end{tabularx}
\end{table}

The 7th LSVOS challenge at ICCV 2025 features two distinct tracks: a Video Object Segmentation (VOS) track evaluated on the complex LVOS and MOSE datasets~\cite{MOSE,MOSEv2}, and a Referring Video Object Segmentation (RVOS) track utilizing the motion-focused MeViS benchmark~\cite{ding2023mevis}.
As shown in \cref{tab:challenge}, our method achieved third place in the competition on the RVOS track, demonstrating strong performance in addressing these challenging scenarios.
We achieve this result by using 12 frames during inference.

\section{Conclusion}
\label{sec:conclusion}
In this work, we demonstrate that the widely adopted, state-of-the-art model Sa2VA suffers from inconsistencies between training and inference, which harm the performance.
To resolve this, we propose Sa2VA-i, an improved version of Sa2VA for which the inference procedure is amended to be consistent with the training procedure.
Sa2VA-i makes initial mask predictions without memory components during both training and inference, preventing the incompatibility between the finetuned mask decoder and frozen memory operations that occurs for Sa2VA.
To make predictions on full videos, Sa2VA-i employs the original, frozen SAM2 model weights to propagate the initial mask predictions across all video frames, only slightly increasing the required memory.
Moreover, Sa2VA-i applies uniform frame sampling for better compatibility with the offline referring video segmentation task.
Together, these changes yield significant improvements, causing Sa2VA-i to considerably outperform Sa2VA and obtain new state-of-the-art results.
Our findings highlight the importance of ensuring consistency between training and inference, and matching the data sampling strategy to the task at hand.

\PAR{Acknowledgments.}\hspace{-5pt}
A. Nekrasov acknowledges funding by BMBF project ``WestAI" (grant no. 01IS22094D).
Compute resources were granted by the Gauss Centre for Supercomputing e.V. through the John von Neumann Institute for Computing on the GCS Supercomputer JUWELS at Jülich Supercomputing Centre.

{
    \small
    \bibliographystyle{ieeenat_fullname}
    \bibliography{main}
}

\end{document}